# Fine-Grained Sentiment Analysis of Electric Vehicle User Reviews: A Bidirectional LSTM Approach to Capturing Emotional Intensity in Chinese Text

Shuhao Chen, Chengyi Tu

# Introduction

As the electric vehicle (EV) industry and its associated charging infrastructure continue to expand rapidly, user reviews have become a crucial resource for evaluating product performance, user experience, and the effectiveness of charging solutions[1]. However, traditional sentiment analysis models typically rely on simplistic binary classifications (e.g., "positive" or "negative") or multi-class categorizations (e.g., "positive," "neutral," and "negative")[2]. While these methods may be adequate in straightforward contexts, they often fail to capture the complexity of emotional expression and the varying degrees of emotional intensity found in user reviews[3].

Furthermore, user reviews frequently reflect multi-dimensional emotional responses, with users expressing different levels of intensity for various aspects of the same product. For example, a reviewer may express strong satisfaction with a vehicle's design while simultaneously highlighting dissatisfaction with its range or charging speed[4]. Traditional sentiment classification approaches tend to oversimplify these nuanced emotions into rigid categories, such as "positive" or "negative," thus failing to accurately represent the full spectrum of emotional intensity and complexity. In contrast, sentiment scoring models offer a more refined approach by assigning quantitative scores to user sentiment, enabling a more detailed understanding of user experiences.

By adopting sentiment scoring models, automotive and charging infrastructure providers can conduct more granular analyses of user feedback, allowing them to pinpoint specific strengths and weaknesses of their products and implement more targeted improvements in both product design and service offerings[5]. As a result, sentiment scoring models are emerging as a more advanced and effective method for sentiment analysis.

This study proposes the development of a sentiment scoring model based on Bidirectional Long Short-Term Memory (Bi-LSTM) networks to analyze the emotional intensity of user reviews related to EV charging infrastructure[3]. Unlike traditional sentiment classification models, the proposed model assigns a sentiment score ranging from 0 to 5 to each review, thus providing a more precise quantification of emotional expression. This approach aims to capture the subtleties of user sentiment, offering a more accurate and meaningful analysis of user experiences.

# Method

## Data preparation

Data processing plays a critical role in enhancing model performance and ensuring high training quality. This study utilizes user review data from PC Auto, a prominent Chinese platform dedicated to evaluating charging pile performance[6]. A comprehensive suite of processing steps, including data cleansing, tokenization, stop word removal, and serialization, is employed to structure and normalize the input data, thereby aligning it with the requirements of deep learning models.

### Data Source

The dataset comprises user-generated reviews on PC Auto, which evaluate various aspects. These reviews form a foundational basis for sentiment analysis and are supplemented by user-provided ratings (on a scale of 0 to 5), serving as labels for supervised learning. A total of 43,678 valid reviews are utilized as training and validation samples for the model, ensuring a robust dataset for analysis.

### Data Cleaning

Before processing, the raw data undergoes meticulous cleaning to eliminate noise that could negatively impact model training. The primary objective of these cleaning processes is to ensure the cleanliness and relevance of the input data, thereby minimizing noise and enhancing the model's accuracy and generalization capabilities[7]. The specific steps involved in data cleaning include:

1. Removing HTML Tags: Reviews often contain embedded HTML code, which is systematically removed to preserve only the relevant textual content.

2. Eliminating Special Characters: Reviews may include extraneous elements such as emojis, punctuation marks, and other non-alphanumeric characters. These elements are excluded, as they do not contribute meaningfully to sentiment analysis.

3. Deduplication: To ensure data integrity, duplicate reviews submitted by the same user are identified and removed from the dataset.

### Data Preprocessing

The primary aim of data preprocessing is to transform the cleaned text data into a structured format that is suitable for model input. Key steps in this phase include tokenization, stop word removal, serialization, and padding/truncation.

1. Tokenization[8]: Unlike English, Chinese text lacks natural delimiters such as spaces, necessitating the use of tokenization. The widely adopted Jieba tool is employed to segment sentences into word sequences, as this word-based approach better captures the semantics of the text in natural language processing. Tokenization allows the model to conduct sentiment analysis based on word sequences rather than individual characters, thus enhancing semantic comprehension.

2. Stop Word Removal[9]: In natural language processing, stop words—high-frequency words such as "的," "是," and "在"—contribute little semantic value. While these words may aid in sentence structure, they do not facilitate sentiment analysis and can introduce unnecessary noise. By removing stop words, redundancy in the data is reduced, enabling the model to focus on sentiment-relevant vocabulary. The Baidu stop word list is utilized for this purpose.

3. Serialization[10]: The tokenized text data requires conversion into a numerical format that is suitable for model processing. A tokenizer is employed to map each word to a unique integer value based on its frequency, with more common words receiving smaller IDs and rarer words larger IDs. This transformation enables the representation of text data as integer sequences for model input. The chosen tokenizer is particularly advantageous for Chinese due to its flexibility in accommodating the language's unique characteristics. It dynamically adjusts vocabulary based on word frequency, controls vocabulary size to mitigate the impact of rare words, and preserves the sequential information of words to effectively capture contextual semantics.

4. Padding and Truncation: Long Short-Term Memory (LSTM) models necessitate input sequences of consistent lengths, while the lengths of actual reviews may vary. To achieve uniformity, reviews shorter than 100 words are padded with zeros, and those exceeding this length are truncated[11]. This approach ensures consistent input lengths, facilitating effective processing by the LSTM.

5. Label Normalization: Given that model outputs typically range between 0 and 1, while sentiment ratings span from 0 to 5, it is essential to normalize the original rating labels prior to training[12]. This is accomplished by dividing the ratings by 5, aligning the label range with the model outputs.

## Vocabulary Coverage Analysis

Vocabulary coverage analysis is conducted to determine an appropriate vocabulary size that maximizes coverage while minimizing the loss of textual information. By calculating coverage rates based on word frequency, we assess the proportion of review texts encompassed by varying vocabulary sizes[13]. The results reveal that 7,119 words achieve 95% coverage, while 16,403 words attain 98% coverage (see Fig. 1). Balancing computational efficiency and model complexity, a vocabulary size of 7,119 words is ultimately selected to ensure substantial retention of critical information while mitigating the adverse effects associated with low-frequency words on model performance.

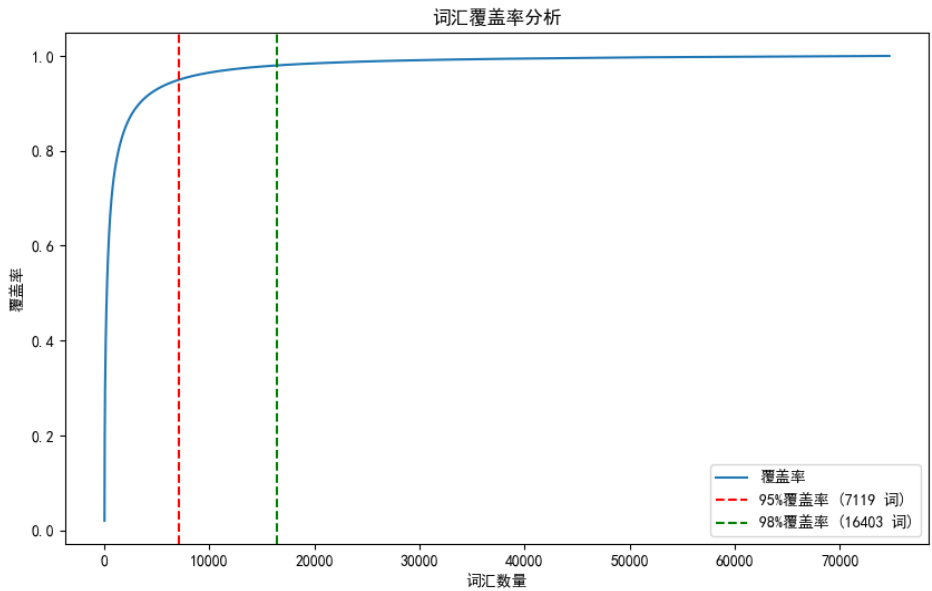

Figure 1: Vocabulary coverage rate as a function of vocabulary number. The red and green line represent the 95% and 98% coverage threshold, respectively.

## Model Construction

The BiLSTM model is an extensively adopted neural network architecture in the domain of sentiment analysis. Distinct from conventional unidirectional LSTMs, BiLSTMs are adept at capturing contextual information in both forward and backward directions simultaneously[14]. This capability is particularly vital in sentiment analysis, where emotional expression is often heavily influenced by surrounding context. In natural language processing, the sentiment conveyed in a sentence frequently relies not only on the preceding words but also on those that follow. For example, in evaluations of charging piles, one might encounter a statement like "The charging pile is very convenient," which could conclude with a contrasting remark such as "but the charging speed is average." If the analysis were to depend solely on forward information, the model could potentially overestimate the positive sentiment. By employing a BiLSTM, the model benefits from a comprehensive analysis of the sentence's context, enabling it to yield more precise sentiment predictions[15].

## Model Architecture Design

In the development of a sentiment analysis model leveraging BiLSTM networks[15], a hierarchical neural network architecture is meticulously designed to process user-generated reviews and predict corresponding sentiment scores. This architecture encompasses following integral components, each fulfilling a distinct function at various stages of data processing:

Figure 2: Architecture of the model architecture (TBD).

1. Embedding Layer: The model's input consists of integer sequences derived from a tokenizer, where each integer symbolizes a specific word within the review. This layer outputs a fixed-dimensional vector representation (word embedding) for each word, utilizing 128-dimensional embeddings to encapsulate semantic relationships, whereby semantically analogous words are positioned closer together in the vector space. By converting textual data into these semantic vectors, the model is equipped to learn inter-word semantic relationships, thereby enhancing the accuracy of sentiment analysis.

2. BiLSTM Layer: The input to this layer is the sequence of word embeddings. The BiLSTM processes the text in both forward and backward directions, thereby extracting emotional features embedded within the reviews. This dual-directional processing captures contextual dependencies more effectively, with 52 LSTM units configured to extract high-level features from the reviews. Such a comprehensive approach is particularly advantageous for handling complex and lengthy Chinese reviews, facilitating a more precise identification of sentiment orientation.

3. Dropout Layer: To mitigate the risk of overfitting, especially pertinent in scenarios involving limited datasets, a Dropout layer is strategically incorporated following the LSTM layer. This mechanism randomly omits connections between neurons, thus reducing the model's dependency on specific neurons and enhancing its generalization capabilities. Bayesian optimization has determined an optimal Dropout rate of 0.007038, introducing randomness that bolsters the model's robustness when confronted with previously unseen review data.

4. Dense Layer: The Dense layer's output translates the features extracted by the LSTM layer into a sentiment score. Utilizing a single neuron with a linear activation function, this layer produces a real-valued score that reflects the model's predicted sentiment rating within a range of 0 to 5. By aggregating the features, this layer ensures that the output accommodates continuous values, rendering it suitable for regression tasks.

5. Loss Function and Optimizer: The Mean Squared Error (MSE) is employed as the loss function, an optimal choice for regression tasks as it effectively quantifies the discrepancy between predicted and actual scores, thereby directing the model's optimization efforts toward enhanced accuracy. The Adam optimizer is utilized, combining momentum with adaptive learning rates to facilitate rapid convergence while minimizing the potential for entrapment in local optima.

In summary, the proposed model architecture is carefully designed to address the specific demands of sentiment analysis in Chinese text. The Embedding layer provides rich semantic representations, while the BiLSTM layer effectively captures contextual dependencies within the data. The inclusion of the Dropout layer mitigates the risk of overfitting, enhancing the model's generalization. The Dense layer generates sentiment scores, with the Mean Squared Error (MSE) serving as the loss function and the Adam optimizer refining the training process[16]. Together, these components demonstrate strong learning and generalization capacities, with each layer playing a critical role in accurately extracting sentiment scores from the input textual data.

# Hyperparameter Optimization

The selection of hyperparameters in deep learning models is critical for achieving optimal performance outcomes. In contrast to learnable parameters, such as weights that are adjusted during training, hyperparameters must be specified prior to the training phase, necessitating careful consideration[17]. The choice of hyperparameters can substantially affect both the efficiency of training and the model's generalization capability to unseen data. Commonly tuned hyperparameters include the learning rate, the number of hidden units, batch size, and various regularization terms. Accurate selection of these hyperparameters is essential to ensure effective learning and to mitigate the risks of overfitting and underfitting, both of which can degrade model performance. To further optimize performance and reduce computational time, hyperparameter optimization techniques are employed to identify the most effective configurations. Arbitrary selection or reliance on default hyperparameters may result in suboptimal performance or even failure in the training process. Therefore, a systematic approach to hyperparameter optimization is necessary to significantly improve the model's efficacy and robustness.

In this study, we employ Bayesian Optimization[18], a global optimization technique, to determine optimal hyperparameters for the model, specifically focusing on the learning rate, number of LSTM units, and dropout rate. This approach is particularly well-suited for optimizing complex, computationally intensive functions, as frequently encountered in deep learning applications. Unlike traditional grid or random search methods, Bayesian Optimization utilizes information from prior evaluations to inform the selection of hyperparameters, providing several distinct advantages. Therefore, this approach not only improves model performance but also enhances the efficiency of the optimization process.

1. Learning Rate[19]: The learning rate is a crucial hyperparameter that controls the magnitude of weight updates during training. A learning rate that is too small may result in slow convergence, while a rate that is too large may cause instability and divergence. For this study, we define the optimization range for the learning rate between $10^{-4}$ and $10^{-2}$, allowing a thorough exploration of potential values that could lead to optimal training outcomes.

2. Number of LSTM Units[20]: The number of units in each LSTM layer corresponds to the capacity of the model to capture temporal features in the data. While increasing the number of LSTM units can enhance the model's ability to represent complex temporal dependencies, it also increases computational demand and the risk of overfitting. We set the optimization range for the number of LSTM units between 32 and 128, enabling a balance between model complexity and computational efficiency.

3. Dropout Rate[21]: Dropout is a regularization technique used to mitigate overfitting by randomly deactivating a subset of neurons during training, thereby promoting better generalization. For this study, we explore dropout rates within the range of 0.2 to 0.6, allowing for a comprehensive evaluation of its impact on model performance.

# Result

## Experimental Procedure

The experimental process begins with an exploratory hyperparameter search, followed by refinement through Bayesian Optimization based on the initial outcomes. The overall procedure was systematically divided into the following phases:

1. Initial Exploration: A broad, random search is conducted to sample various hyperparameter configurations. The model is trained with these configurations, and performance is evaluated on the validation set. This exploratory phase provides essential prior information to inform the subsequent Bayesian Optimization process.

2. Bayesian Optimization: Using the results from the initial phase, Bayesian Optimization is implemented to iteratively refine the hyperparameter configurations. The optimization is guided by the objective function—specifically, the MAE on the validation set—leading to a progressive convergence toward optimal hyperparameters. Each iteration produces significant reductions in the validation error, underscoring the effectiveness of Bayesian Optimization in improving model performance.

3. Experimental Results: After nine optimization iterations, the validation error reaches a stable point, resulting in the following optimized hyperparameter values: Learning Rate: 0.005358, LSTM Units: 52, Dropout Rate: 0.007038. Following optimization, the model exhibits a notable reduction in MAE on the validation set, reflecting enhanced robustness. Additionally, the optimized hyperparameters contribute to improved model stability and accelerated convergence during the training process.

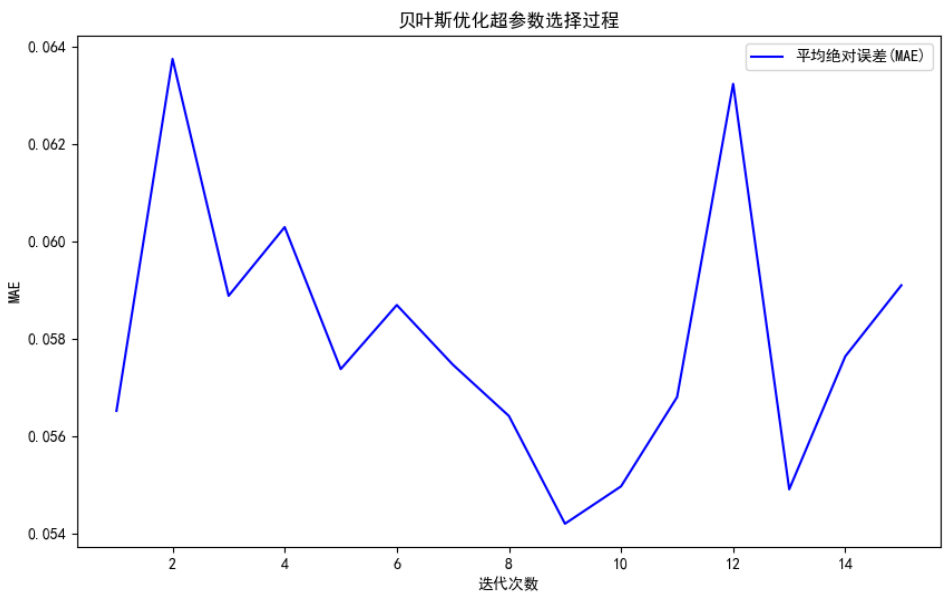

Figure 3: Mean Absolute Error (MAE) as a function of the number of iterations during the Bayesian Optimization process.

The progression of MAE over successive iterations of Bayesian Optimization is depicted in Fig. 3, illustrating a consistent reduction in MAE, particularly during the initial stages of the process. A notable, albeit temporary, increase in MAE occurs at the twelfth iteration, which can be attributed to the inherent exploratory nature of Bayesian Optimization. This exploratory phase often involves evaluating hyperparameter configurations that may not immediately yield performance improvements, thereby preventing premature convergence to suboptimal local minima[22]. This transient rise in MAE underscores the importance of exploration in ensuring a thorough search across the hyperparameter space, which ultimately enables the identification of optimal configurations. Following this brief fluctuation, a marked decline in MAE, followed by stabilization, is observed, signaling the successful discovery of an optimal hyperparameter set. Overall, this analysis validates the efficacy of the Bayesian Optimization process, highlighting its ability to navigate complex, high-dimensional hyperparameter landscapes. Despite occasional performance variations in specific iterations, the optimization process reliably converges toward a robust and effective solution.

## Model Training

After determining the optimal hyperparameter configuration, the final training phase of the model is initiated. The hyperparameters—learning rate, number of LSTM units, and dropout rate—are precisely tuned using Bayesian optimization. The model is trained over 100 iterations, with performance evaluations conducted on a validation set after each iteration. This iterative evaluation enables continuous monitoring of validation error fluctuations, minimizing the risk of overfitting. The training configuration is detailed as follows:

1. Optimizer: Optimizer: The Adam optimizer is utilized, selected for its ability to combine momentum with adaptive learning rates. This optimizer enhances convergence speed, particularly in complex neural network architectures, by adjusting learning rates during training.

2. Loss Function[23]: Mean Squared Error (MSE) is employed as the loss function due to its effectiveness in quantifying the difference between predicted and actual sentiment scores. MSE's sensitivity to larger errors, as it sums squared deviations, renders it well-suited for tasks requiring precision in error measurement.

3. Training and Validation Set Partitioning: To improve the model's generalization capability, the dataset is split, with 80% designated for training and 20% reserved for validation[24]. Performance is evaluated on the validation set after each epoch, providing continuous feedback on the model's generalization and overall efficacy.

Throughout the training process, both training and validation errors are rigorously monitored. This ensures that the model not only performs well on the training data but also demonstrates robust generalization to unseen data, thereby confirming its predictive accuracy and reliability.

# Evaluation Metrics

Upon completing the training phase, the model's performance is rigorously evaluated using two primary metrics:

1. Mean Squared Error (MSE)[25]: Serving as the model's principal loss function, MSE measures the squared differences between predicted and actual sentiment scores. Its sensitivity to larger errors makes it particularly effective in identifying significant deviations, thus providing crucial insights into the model's predictive precision, especially in instances of extreme variance between predictions and true values.

2. Mean Absolute Error (MAE): Unlike MSE, MAE offers a more interpretable evaluation by averaging the absolute differences between predicted and actual values[26]. This metric provides a straightforward assessment of the model's practical accuracy. Given that sentiment scores are bounded within a 0-5 range, a lower MAE signifies a closer alignment between predictions and real-world outcomes, thereby underscoring the model's applicability and reliability in practical scenarios.

# Training Results and Analysis

The model's performance across both training and validation datasets is systematically summarized as follows:

1. Training Error: A detailed analysis of the loss curve throughout the training process indicates a consistent reduction in training error, corresponding with an increasing number of iterations. The error ultimately stabilizes after several iterations, demonstrating the model's ability to effectively capture and internalize the underlying patterns present in the training data.

2. Validation Error: The validation error curve exhibits a sharp initial decline during the early stages of training, followed by a stabilization phase characterized by minor fluctuations. This trend suggests a lack of significant overfitting. After 100 iterations, the MAE for the validation set reached a plateau at a low level, indicating the model's strong predictive performance on previously unseen data.

A comparative assessment of the training and validation errors yields several key insights. The model demonstrates a robust generalization capability, as evidenced by the minimal disparity between the training and validation errors, indicating a high resistance to overfitting. Furthermore, the model continuously integrates new features with each iteration, as reflected by the steady decrease in both training and validation errors, supporting the conclusion that the model converges effectively.

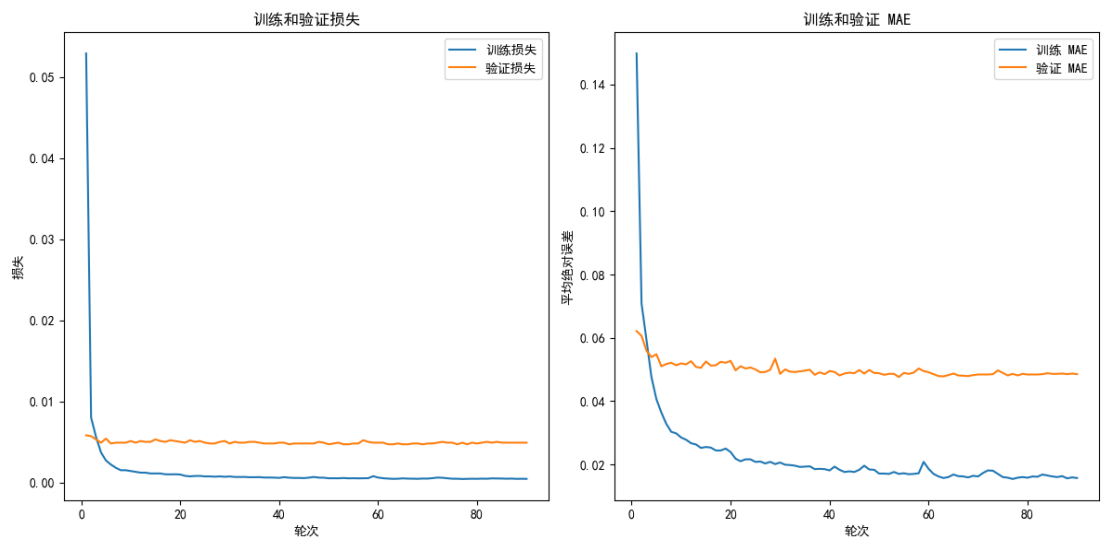

Figure 4: Loss and MAE as a function of iteration number for both training and validation.

In the final phase of experimentation, the model is evaluated using a test dataset, producing performance metrics that validate its applicability in practical settings[27]. The MAE results from the test set further confirm that the optimized model maintains high accuracy and robustness across a range of data scenarios. Figure 4 presents the observed trends in loss and MAE throughout the training process. The left panel illustrates a gradual reduction in the training loss, which stabilizes as training progresses, while the validation loss remains consistently low with minimal fluctuations. This pattern suggests that the model effectively fits the training data without any indication of overfitting. Conversely, the right panel shows the evolution of MAE for both the training and validation sets over the epochs, where the training set MAE steadily decreases, and the validation set MAE stabilizes at a low value. These findings further emphasize the model's robust generalization capabilities, allowing for accurate predictions on unseen data. Overall, the model demonstrates strong performance during both training and validation phases, with stable and low error metrics on the validation set, confirming its reliability and efficacy.

# Discussion

A detailed quantitative comparison between BiLSTM networks and the SnowNLP framework is performed, focusing on their performance across various natural language processing tasks, particularly sentiment analysis[28]. Tab. 1 presents a comprehensive comparison of these models across multiple evaluation metrics.

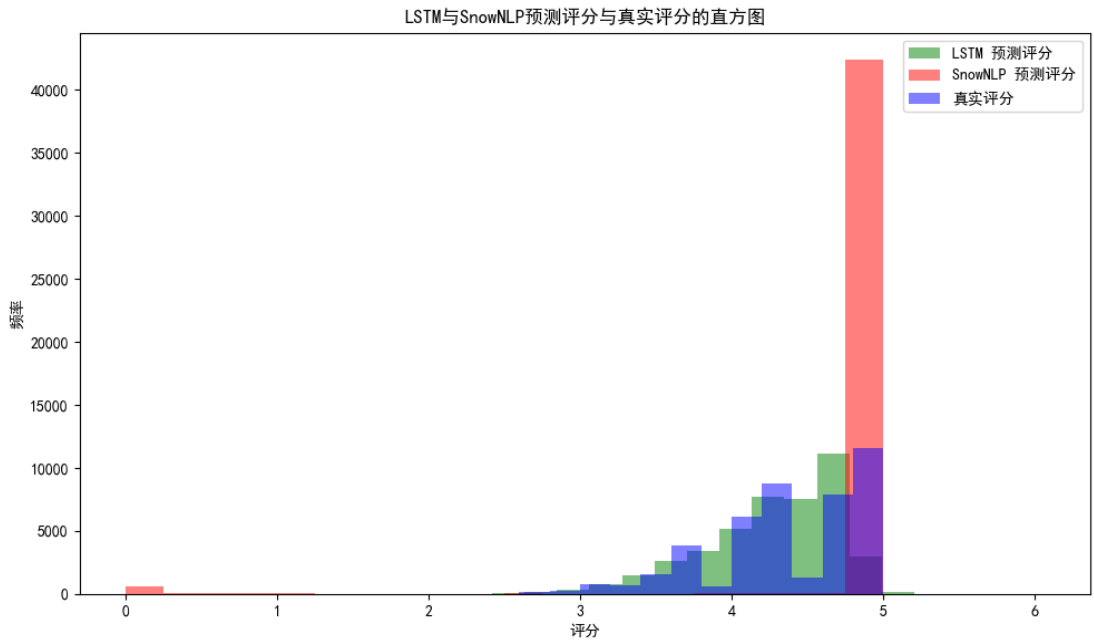

Figure 5. The histograms of BiLSTM and SnowNLP scores in comparison to the actual scores.

Table 1: Comparison of the BiLSTM and SnowNLP models across classic evaluation metrics.

| Evaluation Metrics | BiLSTM | SnowNLP |
|---|---|---|
| Mean Squared Error (MSE) | 0.0958 | 0.9683 |
| Root Mean Squared Error (RMSE) | 0.3096 | 0.984 |
| Mean Absolute Error (MAE) | 0.2338 | 0.6924 |
| Mean Absolute Percentage Error (MAPE) | 5.45% | 17.66% |
| Mean Squared Logarithmic Error (MSLE) | 0.0036 | 0.067 |
| Median Absolute Error (MedAE) | 0.1846 | 0.62 |

Mean Squared Error (MSE) and Root Mean Squared Error (RMSE): The BiLSTM model demonstrates significantly lower MSE and RMSE values compared to the SnowNLP model, indicating more accurate predictions and narrower error margins. This performance gap is especially evident in the analysis of longer textual reviews, where the BiLSTM model consistently maintains a lower error rate. These findings underscore its superior capability in managing more complex and extended text data.

Mean Absolute Error (MAE) and Median Absolute Error (MedAE): In terms of MAE and MedAE, the BiLSTM model outperforms SnowNLP, with its errors being approximately one-third of those produced by SnowNLP[29]. This result suggests that the BiLSTM model excels in capturing the intensity of sentiments in reviews, introducing less predictive bias. Such performance highlights its robustness for fine-grained sentiment analysis tasks.

Coefficient of Determination ($R^2$) and Explained Variance Score (EVS): The BiLSTM model achieves an $R^2$ value of 0.6430 and an EVS of 0.6789, indicating that it accounts for approximately 64% of the variance in the dataset. These metrics point to its strong explanatory power. Conversely, the SnowNLP model yields negative $R^2$ and EVS values, signifying its inability to capture the underlying variance in sentiment fluctuations, thus emphasizing its limitations in this context.

Mean Absolute Percentage Error (MAPE) and Mean Squared Logarithmic Error (MSLE): The MAPE of the BiLSTM model stands at 5.45%, which is markedly lower than SnowNLP's 17.66%, reinforcing the superior accuracy of the BiLSTM in sentiment analysis. Moreover, the BiLSTM's MSLE is just 0.0036, further demonstrating its effectiveness in minimizing logarithmic error when compared to SnowNLP.

Distribution Analysis: Histogram analysis further validates the superior performance of the BiLSTM model in sentiment prediction[30]. The distribution of predicted sentiment scores from the BiLSTM model is smoother, particularly in the 4-5 score range, closely aligning with the actual sentiment score distribution. In contrast, SnowNLP exhibits a notable clustering of predicted scores around 5, reflecting its inability to differentiate sentiment intensity effectively, resulting in less granularity and greater deviation from actual score distributions. Thus, the BiLSTM model shows greater aptitude in capturing nuanced sentiment variations, leading to more accurate and reliable predictions.

In summary, across all evaluated metrics, the BiLSTM model consistently outperforms SnowNLP in the domain of Chinese sentiment analysis[31], particularly regarding error reduction and variance explanation. These results suggest that the BiLSTM model, through its deep learning architecture, is better suited for handling complex, fine-grained sentiment analysis tasks. In contrast, the traditional SnowNLP model shows significant limitations, particularly in capturing the intensity and subtleties of sentiment in more intricate textual datasets.